\documentclass[conference, a4paper]{IEEEtran}
\IEEEoverridecommandlockouts
\usepackage{cite}
\usepackage[letterpaper, top=1.92cm, bottom=4.3cm, left=1.92cm, right=1.92cm]{geometry}
\usepackage{amsmath,amssymb,amsfonts}
\usepackage{algorithmic}
\usepackage{graphicx}
\usepackage{textcomp}
\usepackage{xcolor}
\usepackage{color}
\usepackage{marvosym}

\def\BibTeX{{\rm B\kern-.05em{\sc i\kern-.025em b}\kern-.08em
    T\kern-.1667em\lower.7ex\hbox{E}\kern-.125emX}}
\begin{document}
\setlength{\columnsep}{0.25 in}
\title{Scenarios Engineering driven Autonomous Transportation in Open-Pit Mines}



\author{

\IEEEauthorblockN{Siyu Teng\IEEEauthorrefmark{2}\thanks{Siyu Teng and Xuan Li contributed equally to this work.}}
\IEEEauthorblockA{\textit{Hong Kong Baptist University}\\
\textit{\& IRADS, BNU-HKBU UIC,}\\
\textit{Zhuhai, 519087, China}\\
siyuteng@ieee.org}

\and

\IEEEauthorblockN{Xuan Li\IEEEauthorrefmark{2}}
\IEEEauthorblockA{\textit{Peng Cheng Laboratory,}\\
\textit{ Shenzhen, 518000, China}\\
lix05@pcl.ac.cn}

\and

\IEEEauthorblockN{Yuchen Li}
\IEEEauthorblockA{\textit{Hong Kong Baptist University}\\
\textit{\& IRADS, BNU-HKBU UIC,}\\
\textit{Zhuhai, 519087, China}\\
liyuchen2016@hotmail.com}

\and

\IEEEauthorblockN{Lingxi Li}
\IEEEauthorblockA{\textit{Indiana University-Purdue}\\
\textit{University Indianapolis}\\
\textit{Indianapolis, 151111, USA}\\
LL7@iupui.edu }

\and

\IEEEauthorblockN{Yunfeng Ai}
\IEEEauthorblockA{\textit{University of Chinese Academy of Sciences,}\\
\textit{\& Waytous Ltd.,}\\
\textit{Beijing, 100049, China }\\
aiyunfeng@ucas.ac.cn}

\and


\IEEEauthorblockN{Long Chen\textsuperscript{\Letter}\thanks{Long Chen is the corresponding author.}}

\IEEEauthorblockA{\textit{Institute of Automation, Chinese Academy of Sciences,} \\
\textit{ \& Waytous Ltd.,}\\
\textit{Beijing, 100190, China}\\
long.chen@ia.ac.cn}

\thanks{Our work was supported in part by the National Natural Science Foundation of China under Grant 62203250, in part by the Young Elite Scientists Sponsorship Program of China Association of Science and Technology under Grant YESS20210289, in part by the China Postdoctoral Science Foundation under Grant 2020TQ1057 and Grant 2020M682823.}
}

\maketitle

\begin{abstract}
One critical bottleneck that impedes the development and deployment of autonomous transportation in open-pit mines is guaranteed robustness and trustworthiness in prohibitively extreme scenarios. In this research, 
a novel scenarios engineering (SE) methodology for the autonomous mining truck is proposed for open-pit mines. SE increases the trustworthiness and robustness of autonomous trucks from four key components: Scenario Feature Extractor, Intelligence $\&$ Index (I$\&$I), Calibration $\&$ Certification (C$\&$C), and Verification $\&$ Validation (V$\&$V). Scenario feature extractor is a comprehensive pipeline approach that captures complex interactions and latent dependencies in complex mining scenarios. I\&I effectively enhances the quality of the training dataset, thereby establishing a solid foundation for autonomous transportation in mining areas. C\&C is grounded in the intrinsic regulation, capabilities, and contributions of the intelligent systems employed in autonomous transportation to align with traffic participants in the real world and ensure their performance through certification. V\&V process ensures that the autonomous transportation system can be correctly implemented, while validation focuses on evaluating the ability of the well-trained model to operate efficiently in the complex and dynamic conditions of the open-pit mines. This methodology addresses the unique challenges of autonomous transportation in open-pit mining, promoting productivity, safety, and performance in mining operations.
\end{abstract}

\section{Introduction}
Open-pit mining operations play a vital role in the extraction of mineral resources, which serve as essential raw materials for various industries worldwide. Within these operations, mineral transportation represents a significant portion of productive time and labor costs. However, transportation of mining trucks encounters substantial challenges that necessitate the optimization of operational efficiency, worker safety, and risk mitigation \cite{SOS, SoS1, SoS2}. 

In reply, many researchers focus their attention on autonomous transportation in open-pit mining \cite{Dir, OS}. Nevertheless, due to the nature of the unstructured environment, the development and deployment of autonomous transportation is quite slow. \textcolor{black}{Firstly, the occurrence of harsh weather conditions, such as blizzards, sandstorms, and heavy rainfall, greatly impact the perception system \cite{MotionPlanning,fusionplanner}, and frequent misdetections pose significant safety challenges for autonomous mining trucks. \cite{56,52_1}} \textcolor{black}{Secondly, due to the highly homogeneous scenarios of mining scenarios and the scarcity of latent features, traditional convolutional feature extraction methods often struggle to achieve an adequate amount of content from this dataset with unbalanced data distribution for subsequent prediction and computation.} As a result, this limitation can result in unpredictable behavioral commands issued by autonomous trucks. Finally, during transportation, mining trucks frequently encounter interactions with other specialized machines, such as electric shovels and excavators \cite{52_2}. The complex operational conditions associated with these interactions present challenges in establishing an efficient approach for the verification and validation of well-trained models. Consequently, ensuring the trustworthiness and robustness of these models becomes a challenging task \cite{ZhuIntersection}.

To address the challenges encountered in open-pit mine autonomous transportation, this research integrates scenario engineering (SE) with autonomous transportation. By incorporating Scenario Feature Extractor, Intelligence $\&$ Index (I$\&$I), Calibration $\&$ certification (C$\&$C), and Validation $\&$ Verification (V$\&$V) in scenarios engineering (SE), the trustworthiness, robustness, and learning capabilities of the autonomous transportation model can be further improved \cite{SEMain}. The extension of this conference paper is publish in \cite{tivdtpi}.



\begin{figure*}[t]
\centerline{\includegraphics[width=0.72\linewidth]{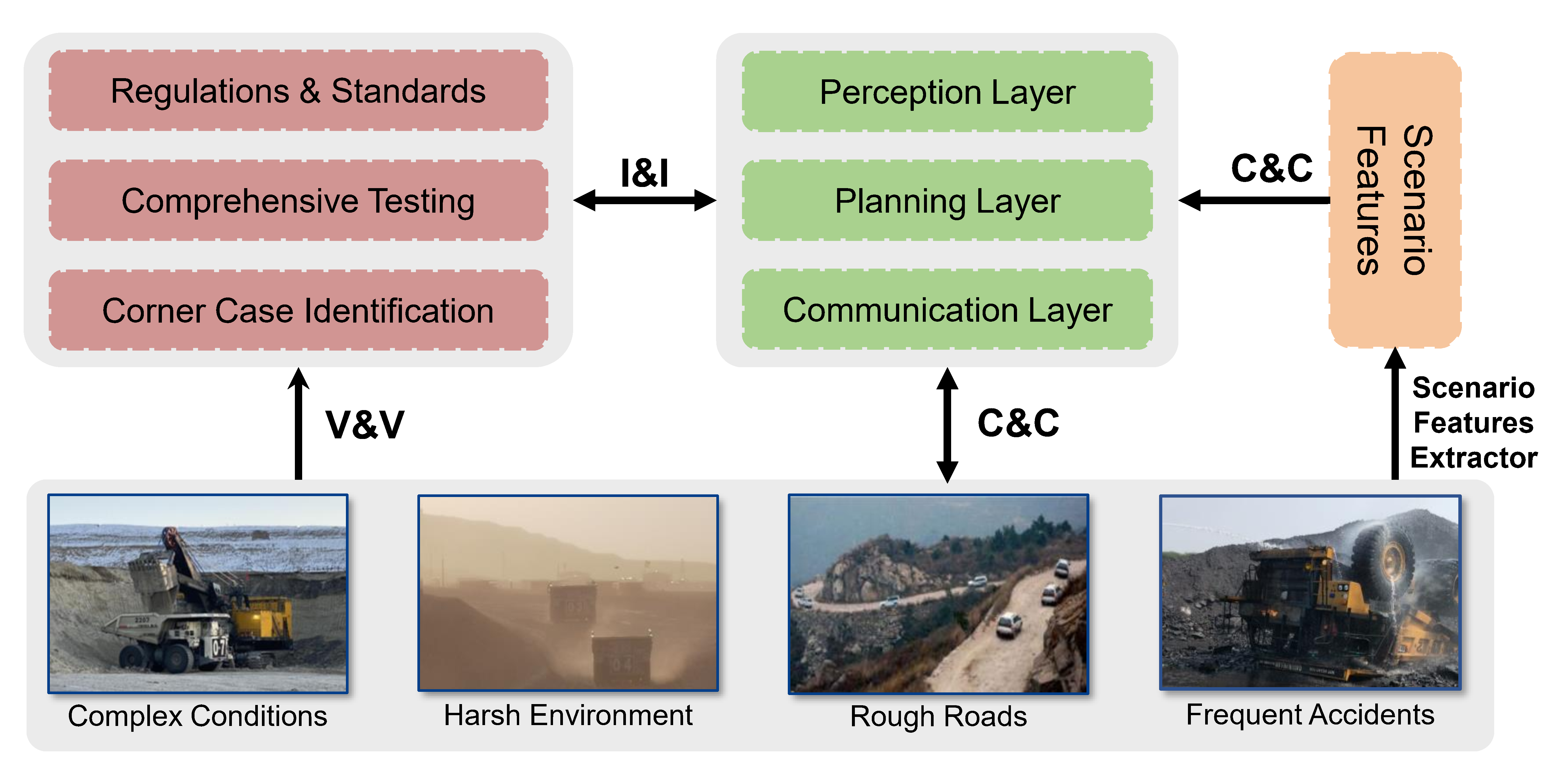}}
\caption{The framework for scenarios engineering (SE) based autonomous transportation in open-pit mining.}
\label{fig: Framework}
\end{figure*}

\section{Scenarios Engineering for Autonomous Transportation}
SE is an emerging scholarly discipline within the field of artificial intelligence (AI) that aggregates the creation, cultivation, and supervision of diverse scenarios to augment the trustworthiness and robustness of AI systems \cite{SE4Metaverse}. This comprehensive methodology involves a profound exploration and meticulous evaluation of AI models with the noble objective of finding potential risks, enhancing learning capabilities, and validating the dependability of trained models. Traditionally, AI development heavily relied on feature engineering \cite{LiDectection}. This approach furnishes a solid foundation for machine learning, but it is criticized for features omitted in the training process and oftentimes proves insufficient in covering the intricate complexity of AI systems and their exposure to various situations \cite{SEMain}. Thus, provides a sufficient complement to features engineering by considering the interplay between AI models and their surroundings, considering the profusion of plausible scenarios they may encounter \cite{SEletter}.
The core of SE lies in ensuring the robustness, trustworthiness, and learning capabilities of AI systems, even in diverse scenarios spanning the real and virtual worlds. These attributes hold paramount significance in the realm of autonomous transportation, especially in the distinctive context of the extremely high demands for safety and efficiency in the transportation of open-pit mines. As shown in Fig. \ref{fig: Framework} illustrates the integration of SE into autonomous transportation in open-pit mines, accomplished through a four-step process aimed at enhancing the feature extraction, robustness, trustworthiness, and learning capabilities of the well-trained model. These steps include Scenario Feature Extractor, Intelligence \& Index, Calibration \& Certification, and Verification \& Validation.

\subsection{Scenario Feature Extractor}
Feature engineering plays a crucial role in establishing a solid fundamental for machine learning. However, it is limited by the reliance on explicit latent feature representation, which limits its adaptability and generalizability \cite{recent1}. One of the limitations of features engineering heavily relies on label features from manually crafted label features, introducing biases and subjectivity that can propagate through the AI system, leading to biased decision-making from human cognitive issues. Additionally, feature engineering struggles with scalability as datasets grow larger and more complex, making it difficult to capture all relevant relationships and update features to accommodate changing requirements. These drawbacks may potentially probability to serious transportation accidents in open-pit mines. In reply, scenarios engineering offers a comprehensive pipeline approach that captures complex interactions and dependencies even within complex mining scenarios, involving various specialized machines such as mining trucks, electric shovels, bulldozers, and excavators \cite{recent}. SE promotes adaptability, generalizability, and fairness by focusing on broader contextual extracting and understanding. It mitigates biases by incorporating features from the entire scenario, enhancing transparency, and providing scalability through automated knowledge techniques \cite{E2eTeng}. 

This approach allows AI systems to learn from multiple facets simultaneously, improving data representation and the ability to address complex issues in both the real and virtual worlds \cite{Miningdataset}. By emphasizing scenarios over fixed labeled features, AI systems developed through scenario engineering exhibit adaptability and generalizability, even when encountering novel situations. Furthermore, scenarios contribute to the reduction of biases by incorporating a broader range of contexts and factors, thereby enhancing fairness, transparency, and accountability. The automated knowledge offered by scenarios engineering facilitates scalability by alleviating the burden of manual feature selection and enabling efficient adaptation to evolving scenarios, which fundamentally addresses the bottleneck imposed by human cognition in neural network learning.


Overall, the adoption of SE empowers the development and deployment of dependable autonomous mining trucks in open-pit mines, enabling them to effectively transport complex and dynamic scenarios. The holistic, adaptable, and automated probability of SE fosters transparency, fairness, and accountability, ultimately leading to the responsible and ethical deployment of autonomous transportation in open-pit mines.

\addtolength{\topmargin}{0.2in}

\subsection{Intelligence $\&$ Index}
SE is supported by six foundational components known as ``6I": cognitive intelligence, parallel intelligence \cite{parallelplanning}, crypto intelligence, federated intelligence, social intelligence, and ecological intelligence. The goal of SE is to achieve six specific goals, referred to as ``6S": safety, security, sustainability, sensitivity, service, and smartness. The intelligence of the infrastructures and the indexes associated with the 6S goals collectively form the concept of Intelligence $\&$ Index (I$\&$I) within SE \cite{SEMain}. 

\begin{figure}[t]
\centerline{\includegraphics[width=0.9\linewidth]{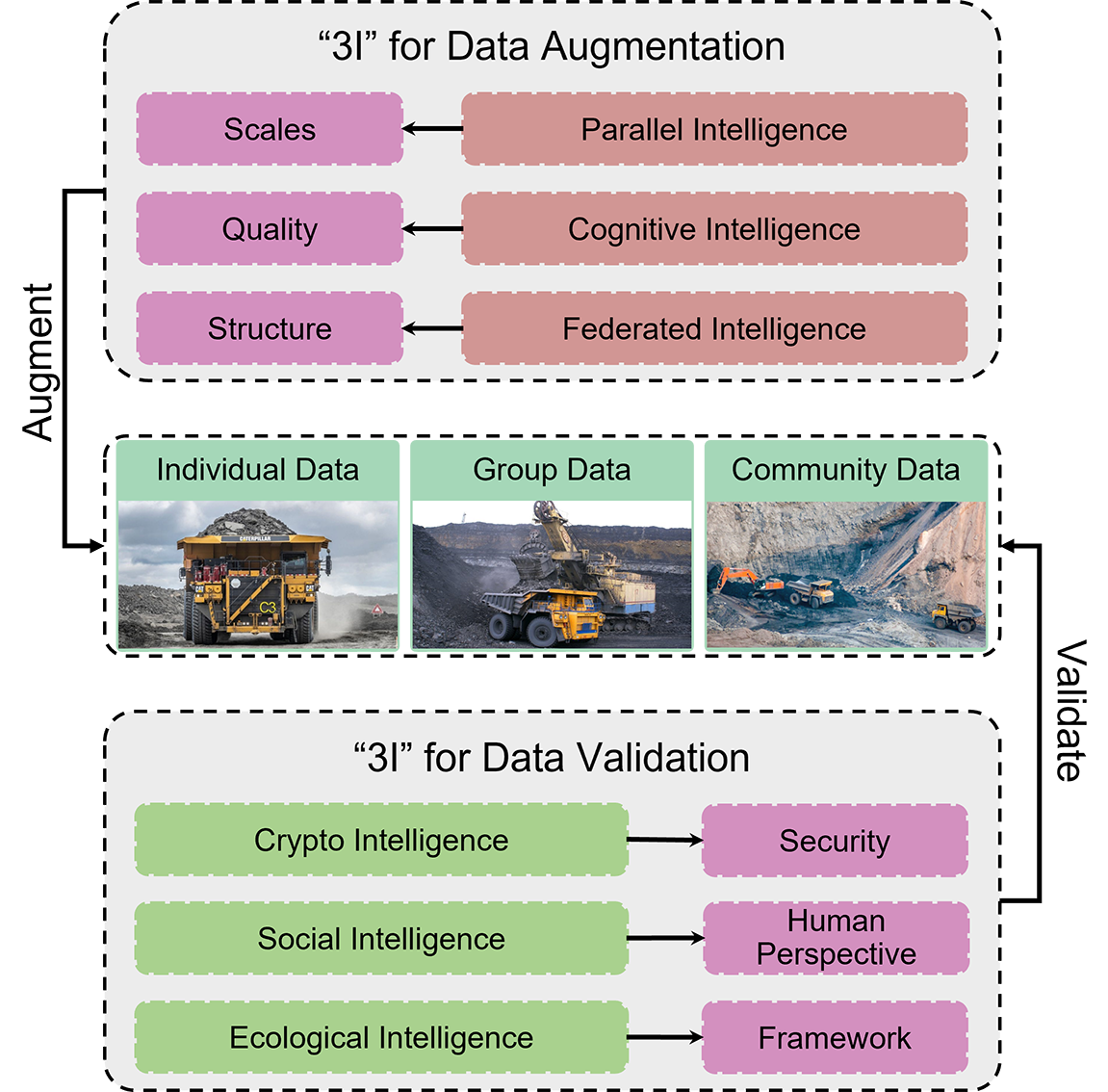}}
\caption{Intelligence $\&$ Index for data argumentation and validation in open-pit mines.}
\label{fig: data}
\end{figure}

In the area of autonomous transportation in open-pit mines, the magnitude and distribution of the dataset limit the ceiling of autonomous transportation models \cite{Li_paralleldataset, Li_paralleldataset2}. To ensure the quality of the dataset employed for model training, as shown in Fig. \ref{fig: data}, the ``3I", including cognitive intelligence, parallel intelligence, and federated intelligence, enhance the dataset in terms of magnitude and distribution, thereby alleviating the limitations related to dataset magnitude and non-uniform distribution. While the remaining ``3I", including crypto intelligence, cognitive intelligence, and ecological intelligence, are employed for data validation. Their primary function is to ensure the trustworthiness of the data, mitigate data bias, and further enhance the overall quality of the dataset. 

\subsubsection{``3I" for data augmentation}

Parallel Intelligence enables data augmentation from scales, which is effective in transforming small data into big data \cite{parallelintelligence, parallellearning}. It can facilitate data augmentation by leveraging artificial systems to model real-world behaviors and conduct computational experiments. By generating synthetic data through parallel intelligence, the original dataset can be augmented. This augmented dataset can be utilized for training and validating AI models, leading to improved performance and generalization.

Cognitive Intelligence enables data augmentation from quality, which focuses on understanding how neural networks process data. It can contribute to data augmentation by enhancing the understanding and representation of data. Algorithms developed through cognitive intelligence can improve data preprocessing, feature extraction, and knowledge representation, thereby augmenting the quality and usefulness of the data.

Federated Intelligence enables data augmentation from structure, which transforms individual data from a single organization into federated data for multiple organizations and divides all data into three levels, individual, group, and community. Federated intelligence enables data federation through federated control and service federation through federated management. It operates under the support and constraints of federated contracts, federated consensus, federated incentives, and federated security.
\subsubsection{``3I" for data validation}

Crypto Intelligence enables data validation from security, which focuses on generating. In the context of data validation, crypto intelligence can contribute by providing a secure and transparent framework for validating and verifying training dataset integrity and authenticity. The use of blockchain technology ensures immutability and verifiability, further enhancing the trustworthiness of validated data.

Social Intelligence enables data validation from the human perspective, which addresses complex decision problems with a focus on social behaviors and relationships. In the context of data validation, social intelligence can play a role in understanding social dynamics related to data generation and validation processes. It can help in identifying biases, social influences, and contextual factors that may impact the validity of the data being validated. By considering the social aspects, social intelligence can contribute to improving the quality and reliability of data validation processes.

Ecological Intelligence enables data validation from structure, which adopts an ecological perspective to solve complex tasks in intelligent systems. It emphasizes the opposition and unity of individual, group, and community levels of data, the integration of natural, social, and cyber environments, and the investigation of deep models, analytics, and management of ecological systems. In the context of data validation, ecological intelligence can provide a holistic approach to ensure the integrity and effectiveness of the validation process. By considering the broader ecological context, including the interdependencies between different data sources, the impact of external factors, and the alignment with ethical and sustainability considerations, ecological intelligence can contribute to robust and comprehensive data validation practices.

In summary, “3I” achieve the goal of data augmentation from scales, quality, and framework, while other ``3I" reache the goal of data validation from security, human perspective, and framework. The implementation of ``6I" effectively enhances the quality of autonomous transportation data, thereby establishing a solid foundation for autonomous transportation in mining areas. In addition, it also promotes the realization of the ultimate goal of ``6S" in mining operations. The concept of ``6I" represents a transformation in AI systems, shifting from feature-based elements, function, and engineering to scenario-based intelligent ecology. This transformation enables the achievement of the ``6S" goals, contributing to the smart development and sustainability of intelligent systems. The 6S stands for safety in the physical world, security in the cyber world, sustainability in the ecological world, sensitivity to individual needs, service for all, and overall smartness. To effectively measure and assess each of the ``6S" goals in SE, corresponding indexes must be designed based on the specific applications and functions. These indexes serve to evaluate and express the attainment of each goal, forming another set of ``6I" components: safety index, security index, sustainability index, sensitivity index, service index, and smartness index.

\subsection{Calibration $\&$ Certification}

C$\&$C (Calibration $\&$ Certification) in autonomous transportation involves identifying suitable parameter values for well-trained models to align with traffic participants in the real world and ensuring their performance through certification. In order to ensure efficient learning efficiency of the unmanned transportation model as well as the effective criterion of the model's performance, C\&C provides a novel methodology.

Calibration is manifested as minimizing the distance between model outputs and labeled commands, while certification validates the capabilities of well-trained models. Certificates represent system knowledge and contribution and establish trust in autonomous mining trucks. C$\&$C enhances the accuracy and reliability of autonomous transportation systems by aligning them with real-world conditions and providing validated performance assurance.

To improve autonomous transportation in an open-pit mine using the C$\&$C method, several crucial steps can be implemented. By learning potential representations from Scenario Feature Extractor of the total scenario, calibration of the autonomous transportation model is crucial. Calibration involves determining optimal values for model parameters, ensuring that the internal dynamics of the AI system closely align with work conditions in the real world. This calibrating process can be achieved by minimizing the aggregate distance between the outputs of the well-trained model and the labeled commands, derived from well-designed scenarios based on well-designed scenarios specific to the open-pit mine environment.

To validate and confirm the technical performance of the calibrated system, certification should be obtained. This confirmation process necessitates obtaining certification, which serves as a formal certification of the characteristics of the autonomous transportation system. The certification must be issued by a reputable third-party organization. To fulfill the certification requirements, the establishment of the following two criteria is essential:

\begin{itemize}
  \item [$\bullet$] 
    Preservation: Certification should be non-perishable and easily preserved, especially for systems designed to operate over long periods of time. Digital certification, particularly in the form of non-fungible tokens (NFTs) residing in a blockchain, can be an ideal candidate for long-term preservation.
  
  \item [$\bullet$]
    Uniqueness: Certification should be globally unique, enabling clear distinction between different systems of the same class. This ensures that each certified autonomous vehicle in the open-pit mine can be uniquely identified and distinguished.   
\end{itemize}

By fulfilling these established criteria, certifications can be granted to acknowledge compliance and performance in autonomous transportation systems. These certificates can be grounded in the intrinsic knowledge, capabilities, and contributions of the intelligent systems employed in autonomous transportation. They can be manifested through non-transferring reputation endorsements or transferring economic tokens, thereby establishing a trustworthy and robust AI system for autonomous mining trucks operating within the open-pit mine.

\begin{figure}[t]
\centerline{\includegraphics[width=0.9\linewidth]{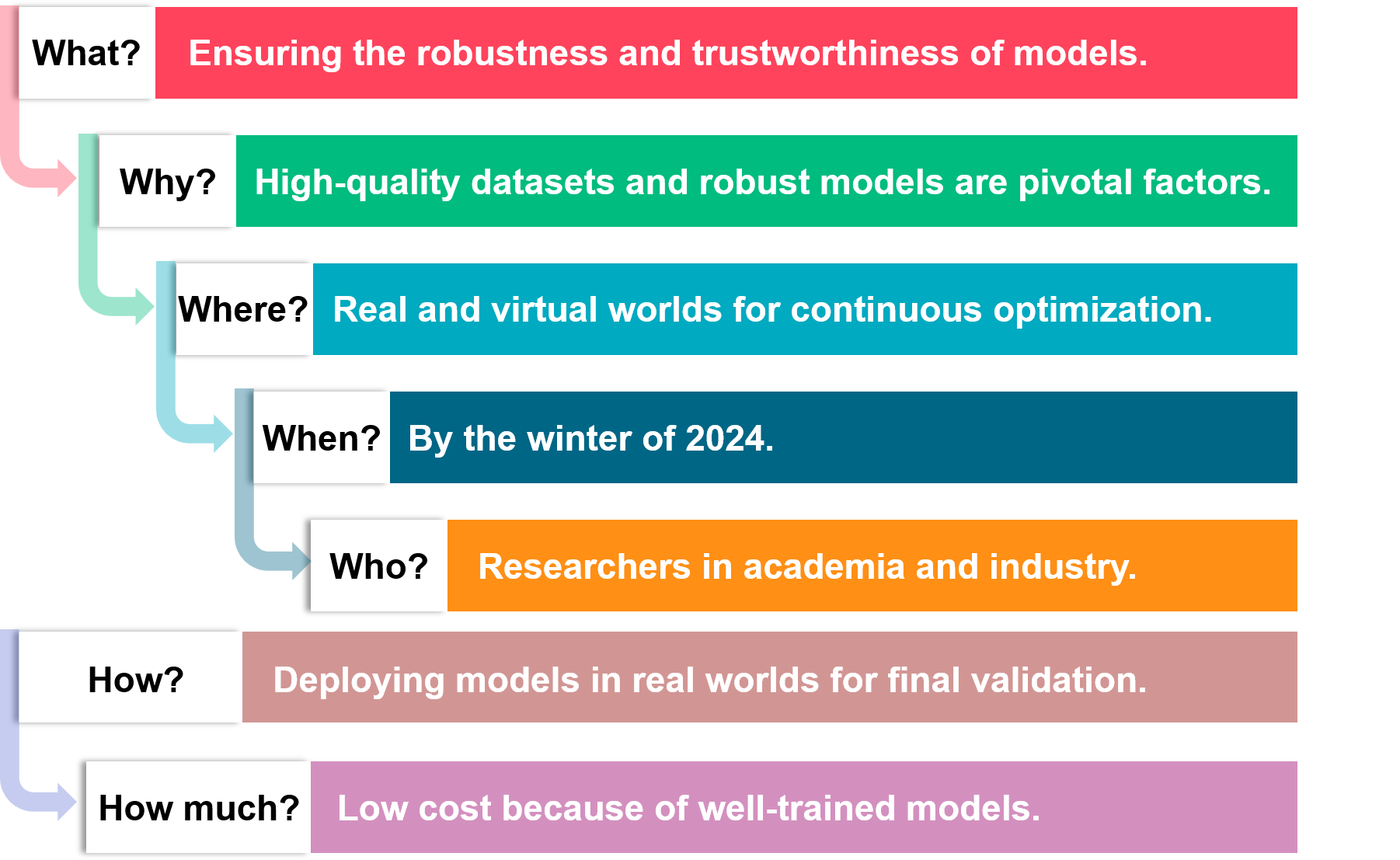}}
\caption{5W2H for autonomous transportation in open-pit mines.}
\label{fig: 5h2w}
\end{figure}
\subsection{Verification $\&$ Validation}

Verification $\&$ Validation (V$\&$V) in autonomous transportation involves ensuring the correctness and performance of the system. Verification focuses on assessing the system's compliance with specifications through activities like design analysis, testing, and inspection. It verifies whether the well-trained model has been implemented correctly and adheres to the specified requirements.

Validation involves evaluating the performance of the autonomous transportation system in real-world environments within open-pit mines. This step aims to validate the functionality of the well-trained model and ensure that it meets the specific requirements and expectations of the end users. Validation includes conducting tests in the actual transport conditions of the open-pit mine and assessing the system's response to various challenges and situations. Any anomalies, gaps, or areas for improvement can be identified, and then, leading to the enhancement of the overall user experience and safety, by validating the system's performance.

To support the V\&V process, the 5W2H analysis can be employed. This analysis gathers relevant information for effective feature extraction from scenario engineering, ensuring that the system's capabilities and performance are appropriately evaluated in various environments. The implementation advice for autonomous transportation in an open-pit mine is illustrated in Fig. \ref{fig: 5h2w}. By following this guidance, the V\&V process can be conducted more effectively, leading to a higher level of trustworthiness and robustness in the autonomous transportation system. Through verification, the implementation of the AI system can be thoroughly examined, while validation ensures that the system functions according to the user-defined requirements.

Due to the strict requirements for physical safety and the unique and demanding nature of the environment in open pit mines, V$\&$V becomes even more critical. The verification process ensures that the autonomous transportation system can be correctly implemented, while validation focuses on evaluating the ability of the well-trained model to operate efficiently in the complex and dynamic conditions of the open-pit mines. Potential risks and issues can be identified, mitigated, and resolved by V$\&$V, ultimately leading to the development of robust and trustworthy autonomous transportation systems that can enhance productivity, safety, and efficiency within open-pit mining environments.

\section{Conclusion and Perspectives}
SE is a comprehensive approach to autonomous transportation for open-pit mines. It incorporates several key components, namely the Scenario Feature Extractor, Intelligence and Index (I\&I), Calibration and Certification (C\&C), and Verification and Validation (V\&V). These components work together to ensure the seamless integration of various system elements, optimize AI model parameters, and certify the overall performance of the autonomous transportation system. Through the rigorous verification of adherence to specifications and validation of functionality in real-world scenarios, SE significantly enhances the robustness, trustworthiness, and effectiveness of autonomous transportation systems. Moreover, this approach is designed to address the unique challenges present in open-pit mining environments, thereby promoting increased productivity, safety, and overall performance in mining transportation. The emergence and utilization of foundation models \cite{Chatgept2} present a novel perspective for advancing the field of SE. Exploring the integration of extensive the intelligence of foundation model and the comprehensive nature of SE stands for an exciting and uncharted research direction.






\bibliographystyle{IEEEtran}
\bibliography{mylib.bib}

\vspace{12pt}

\end{document}